\documentclass{article} % For LaTeX2e
\usepackage[preprint]{colm2026_conference}

\usepackage{microtype}
\usepackage{hyperref}
\usepackage{url}
\usepackage{booktabs}

\usepackage{amsmath}
\usepackage{amssymb}
\usepackage{mathtools}
\usepackage{amsthm}
\usepackage{xcolor}
\definecolor{softblue}{RGB}{40,80,150}  % muted academic blue
\usepackage{multirow}
\usepackage[table]{xcolor}
\usepackage{tcolorbox}

\newtcbox{\hlsecondarytab}{on line, rounded corners, box align=base, colback=green!10, colframe=white,size=fbox,arc=3pt, before upper=\strut, top=-2pt, bottom=-4pt, left=-2pt, right=-2pt, boxrule=0pt}
\newtcbox{\hlprimarytab}{on line, box align=base, colback=red!10,colframe=white,size=fbox,arc=3pt, before upper=\strut, top=-2pt, bottom=-4pt, left=-2pt, right=-2pt, boxrule=0pt}
\newtcbox{\hlgraytab}{on line, rounded corners, box align=base,colframe=white,size=fbox,arc=3pt, before upper=\strut, top=-2pt, bottom=-4pt, left=-2pt, right=-2pt, boxrule=0pt}

\usepackage{wrapfig}

\usepackage{graphicx}
\usepackage{subfigure}

\usepackage{algorithm}
\usepackage{algorithmic}

\usepackage{enumitem}

\usepackage{lineno}

\definecolor{darkblue}{rgb}{0, 0, 0.5}
\hypersetup{colorlinks=true, citecolor=darkblue, linkcolor=darkblue, urlcolor=darkblue}

\title{Experiential Reinforcement Learning}

\author{Taiwei Shi\textsuperscript{1}\thanks{Work done during internship at Microsoft's Office of Applied Research}~, Sihao Chen\textsuperscript{2}, Bowen Jiang\textsuperscript{3*}, Linxin Song\textsuperscript{1}, Longqi Yang\textsuperscript{2}, Jieyu Zhao\textsuperscript{1}\\
\textsuperscript{1}University of Southern California, \textsuperscript{2}Microsoft, \textsuperscript{3}University of Pennsylvania\\
\texttt{\small{\{taiweish@usc.edu, sihaochen@microsoft.com\}}}
}

\usepackage{hyperref}

\usepackage{xargs} 
\usepackage[pdftex,usenames,dvipsnames,table,svgnames]{xcolor}
\usepackage[colorinlistoftodos,prependcaption,textsize=tiny]{todonotes} %[disable]
\usepackage[capitalize,noabbrev]{cleveref}

\theoremstyle{plain}

\theoremstyle{definition}

\theoremstyle{remark}

\usepackage[textsize=tiny]{todonotes}

\begin{document}

\ifcolmsubmission
\linenumbers
\fi

\maketitle

% \begin{document}

% \twocolumn[
% % \icmltitle{Reinforcement Learning from Self-Reflection}
% % \icmltitle{Experiential Reinforcement Learning \\ via Self-Reflection and Experience Internalization}

% % Other candidates
% % Explore, Reflect, and Internalize: \\ Hindsight Reinforcement Learning for Language Models

% % You may provide any keywords that you
% % find helpful for describing your paper; these are used to populate
% % the "keywords" metadata in the PDF but will not be shown in the document

% \vskip 0.3in
% ]




\begin{abstract}
Reinforcement learning has become the central approach for language models (LMs) to learn from environmental reward or feedback. In practice, the environmental feedback is usually sparse and delayed. Learning from such signals is challenging, as LMs must implicitly infer how observed failures should translate into behavioral changes for future iterations. 
We introduce \textit{Experiential Reinforcement Learning} (ERL), a training paradigm that embeds an explicit experience–reflection–consolidation loop into the reinforcement learning process. Given a task, the model generates an initial attempt, receives environmental feedback, and produces a reflection that guides a refined second attempt, whose success is reinforced and internalized into the base policy. This process converts feedback into structured behavioral revision, improving exploration and stabilizing optimization while preserving gains at deployment without additional inference cost. Across sparse-reward control environments and agentic reasoning benchmarks, ERL consistently improves learning efficiency and final performance over strong reinforcement learning baselines, achieving gains of up to +81\% in complex multi-step environments and up to +11\% in tool-using reasoning tasks. These results suggest that integrating explicit self-reflection into policy training provides a practical mechanism for transforming feedback into durable behavioral improvement.
\end{abstract}

\begin{figure*}[h]
    \centering
    \includegraphics[width=\textwidth]{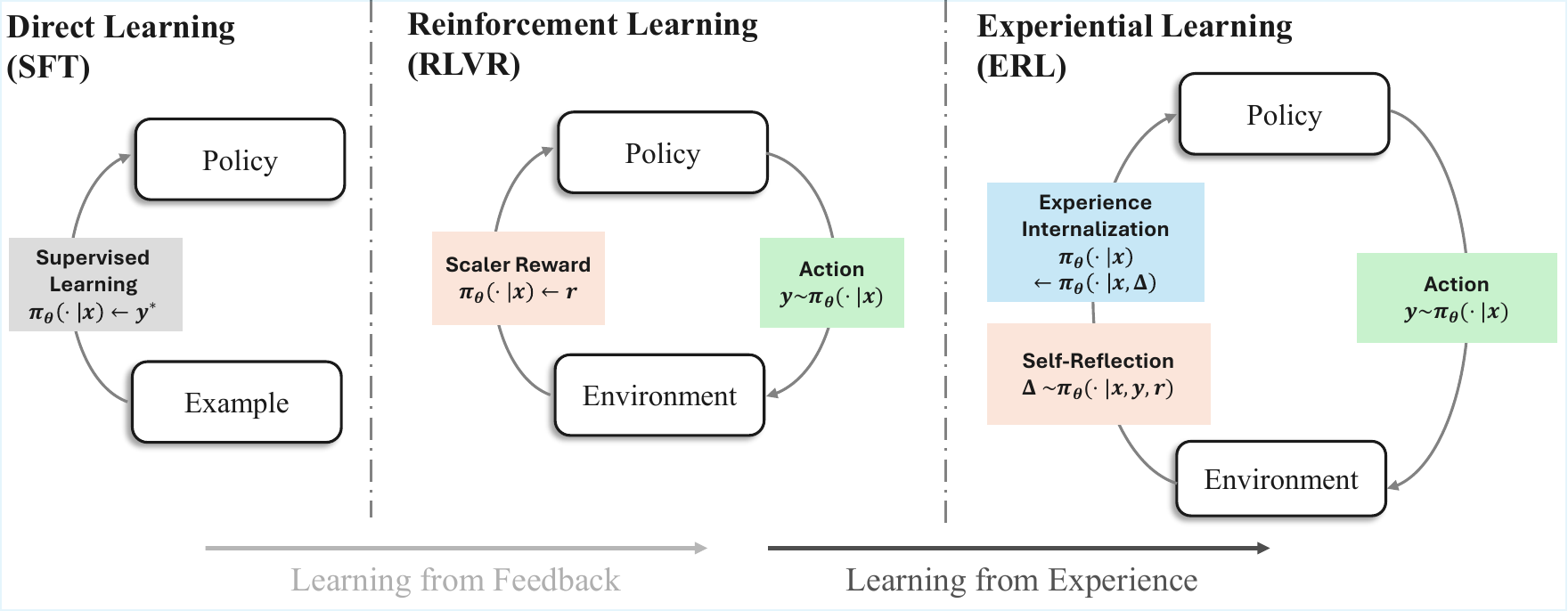}
    \vspace{-1em}
    \caption{
    In \textit{Experiential Reinforcement Learning} (ERL), instead of learning from feedback or outcome directly, an agent learns to (1) verbally reflect on its experience and observed outcome, and (2) internalize the reflections to induce behavioral changes in future iterations. 
    % ERL is inspired by the concept of Experiential Learning \citep{kolb2014experiential} in education.
    }
    \label{fig:rlsr_simple}
\end{figure*}

\vspace{-6pt}
\section{Introduction}

\begin{figure*}[t]
    \centering
    \includegraphics[width=\textwidth]{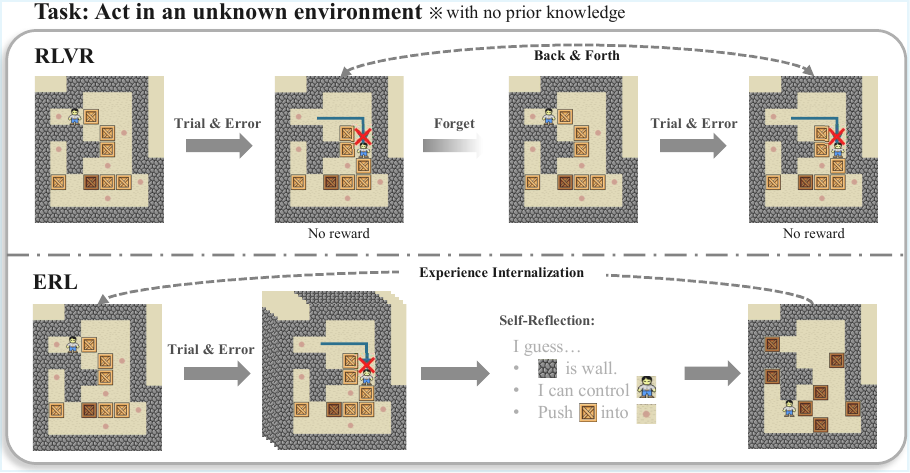}
    \vspace{-1em}
    \caption{Conceptual comparison of learning dynamics in RLVR and Experiential Reinforcement Learning (ERL). RLVR relies on repeated trial-and-error driven by scalar rewards, leading to back-and-forth exploration without durable correction. ERL augments this process with an experience–reflection–consolidation loop that generates a revised attempt and internalizes successful corrections, enabling persistent behavioral improvement.}
    \vspace{-6pt}
    \label{fig:rlsr_comp}
\end{figure*}

Large language models are increasingly deployed as decision-making agents that must act, observe feedback, and adapt their behavior in environments with delayed rewards and partial information \citep{singh2025openaigpt5card, yang2025qwen3technicalreport, song2025coact1computerusingagentscoding, kimiteam2026kimik2openagentic}. Reinforcement learning offers a natural framework for improving such agents. 
The task environments typically provide feedback in the form of outcome reward after an agent generates the entire trajectory. In practice, training agents against such sparse and delayed outcome signals remains difficult, as models must implicitly infer 
% yet practical training remains difficult. In sparse-reward settings, outcome signals arrive late and provide limited guidance about how a policy should change. 
% Even when feedback is available, models must implicitly infer 
how to translate observed failures into corrective behavior, a process that is often unstable and sample-inefficient \citep{zhang2025agentlearningearlyexperience, shi2026efficientreinforcementfinetuningadaptive}. These challenges become more pronounced in agentic reasoning tasks, where multi-step decisions could amplify small errors and obscure credit assignment.

Humans address similar challenges through a process often described as \textit{experiential learning}, in which effective adaptation arises from a cycle of experience, reflection, conceptualization, and experimentation \citep{kolb2014experiential}. After observing an outcome, a learner reflects on what occurred, forms revised internal models, and applies those revisions in subsequent attempts. This cycle transforms raw feedback into actionable behavioral corrections before those corrections are consolidated into future behavior. While language models have demonstrated reflection-like capabilities at inference time, standard reinforcement learning pipelines largely reduce feedback to scalar optimization signals, requiring policies to implicitly discover corrective structure through undirected exploration rather than explicit experiential revision.

% \sihao{In RLVR, we sample $y \sim \pi_\theta(\cdot | x)$ from the policy model $\pi_\theta$ and receives feedback from the environment. The feedback is usually translated into a scaler reward $r$ for policy gradient update on $\pi_\theta$. However }

This perspective highlights a progression in how language models learn from supervision and interaction, illustrated in Figures~\ref{fig:rlsr_simple} and~\ref{fig:rlsr_comp}. In supervised fine-tuning (SFT), policies imitate fixed examples, enabling strong pattern reproduction but offering no mechanism for revising behavior once deployed. Reinforcement learning with verifiable rewards (RLVR) extends learning into interactive settings by optimizing scalar feedback, allowing agents to improve through trial-and-error; however, corrective structure must still be inferred implicitly from sparse or delayed rewards. As visualized in Figure~\ref{fig:rlsr_comp}, this can lead to repeated exploration without durable behavioral correction. A natural next step is to structure learning around experience itself, transforming feedback into intermediate reasoning that supports explicit revision and consolidation within each episode. Figure~\ref{fig:rlsr_simple} conceptualizes this shift as moving from learning purely from feedback toward deliberate learning from experience.

\begin{figure*}[t]
    \centering
    \includegraphics[width=\textwidth]{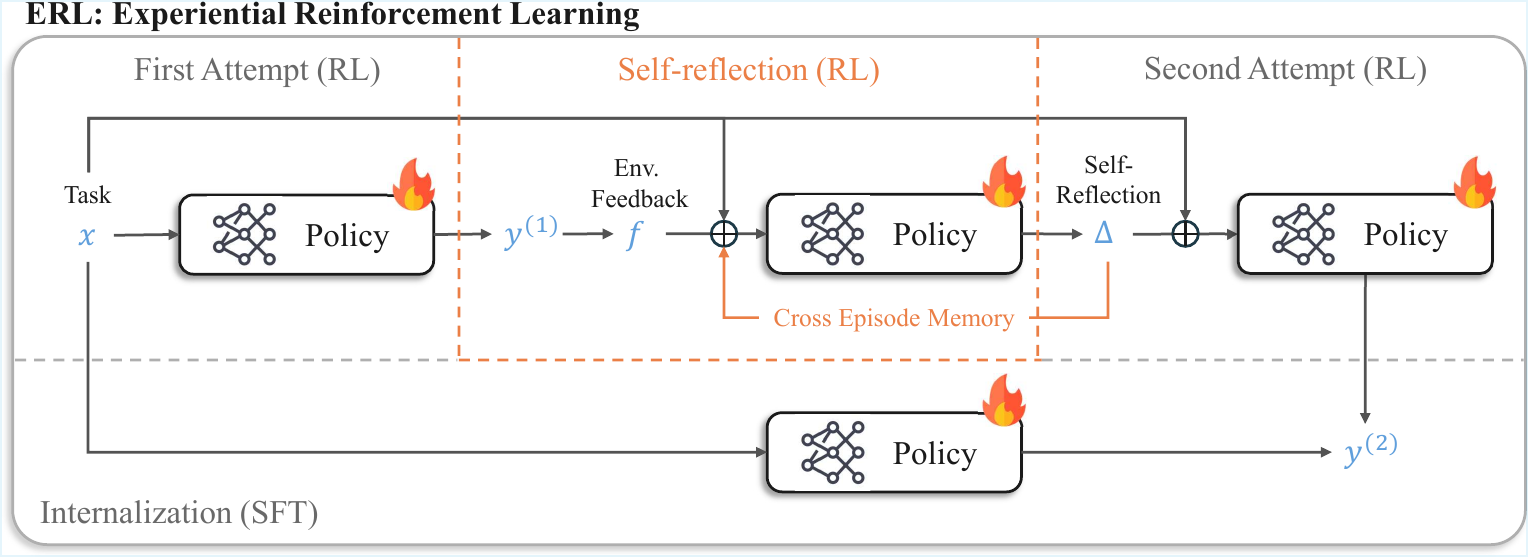}
    \vspace{-1em}
    \caption{
    Overview of Experiential Reinforcement Learning (ERL). 
    Given an input task $x$, the language model first produces an initial attempt and receives environment feedback.
    The same model then generates a self-reflection conditioned on this attempt, which is used to guide a second attempt.
    Both attempts and reflections are optimized with reinforcement learning, while successful second attempts are internalized via self-distillation, so the model learns to reproduce improved behavior directly from the original input without self-reflection. 
    }
    \label{fig:rlsr}
\end{figure*}

In this work, we introduce \textit{Experiential Reinforcement Learning} (ERL), a training paradigm that embeds an explicit experience–reflection–consolidation loop inside reinforcement learning. Instead of learning solely from outcome rewards, the model first produces an initial attempt, receives environment feedback, and generates a structured reflection describing how the attempt should be improved. This reflection conditions a refined second attempt, whose outcome is reinforced and internalized into the base policy. By converting feedback into intermediate reasoning signals, ERL enables the model to perform targeted behavioral correction before policy consolidation. Over time, these corrections become part of the policy itself, allowing improved behavior to persist even when reflection is absent at deployment. An overview of the algorithm is shown in Figure \ref{fig:rlsr}.

We evaluate ERL across sparse-reward control environments and agentic reasoning benchmarks spanning two model scales. ERL consistently outperforms RLVR in all six evaluated settings, achieving gains of up to +81\% in Sokoban, +27\% in FrozenLake, and up to +11\% in HotpotQA. These results demonstrate that embedding structured experiential revision into training improves learning efficiency and produces stronger final policies across both control and reasoning tasks.

\paragraph{Contributions.}
Our main contributions are:

\begin{itemize}[leftmargin=*]
    \item We introduce Experiential Reinforcement Learning (ERL), a reinforcement learning paradigm that incorporates an explicit experience–reflection–consolidation loop, enabling models to transform environment feedback into structured behavioral corrections.

    \item We propose an internalization mechanism that consolidates reflection-driven improvements into the base policy, preserving gains without requiring reflection at inference time.

    \item We demonstrate that experiential reinforcement learning improves training efficiency and final performance across agentic reasoning tasks.
\end{itemize}

% \jz{for fig.1, do we mean we update the policy model across all the steps? Should it be $y_2 \to $ policy model? } \taiwei{yes, we update the model across all steps.}

\section{Experiential Reinforcement Learning (ERL)}
\label{sec:rlsr}

\begin{algorithm}[t]
% \caption{Reinforcement Learning from Self-Reflection}
\caption{Experiential Reinforcement Learning}
\label{alg:rlsr_simple}
\small
\begin{algorithmic}[1]

\STATE \textbf{Inputs:} Language model $\pi_\theta$; dataset of questions $x$; reward threshold $\tau$;
environment returning feedback $f$ and reward $r$.
\STATE \textbf{Initialize:} reflection memory $m \leftarrow \emptyset$.

\REPEAT
    \STATE Sample question $x$ from the dataset. \vspace{5pt}

    \STATE \textcolor{softblue}{\textbf{// First attempt}}
    \STATE Sample an answer $y^{(1)} \sim \pi_\theta(\cdot \mid x)$.
    \STATE Obtain environment feedback and reward $(f^{(1)}, r^{(1)})$. \vspace{5pt}

    \STATE \textcolor{softblue}{\textbf{// Self-reflection}}
    \STATE Sample a reflection $\Delta \sim \pi_\theta(\cdot \mid x, y^{(1)}, f^{(1)}, r^{(1)}, m)$. \vspace{5pt}

    \STATE \textcolor{softblue}{\textbf{// Second attempt}}
    \STATE Sample a refined answer $y^{(2)} \sim \pi_\theta(\cdot \mid x, \Delta)$.
    \STATE Obtain environment feedback and reward $(f^{(2)}, r^{(2)})$.
    \STATE Set reflection reward $\tilde r \leftarrow r^{(2)}$.
    \STATE Store reflection $m \leftarrow \Delta \;\;$ if $\;\; r^{(2)} > \tau$. \vspace{5pt}

    \STATE \textcolor{softblue}{\textbf{// RL update}}
    \STATE Update $\theta$ via
           $\mathcal{L}_{\text{policy}}(\theta)$ over the first attempt, reflection, and second attempt. \vspace{5pt}

    \STATE \textcolor{softblue}{\textbf{// Internalization}}
    \STATE Update $\theta$ via $\mathcal{L}_{\text{distill}}(\theta)$
           to internalize reflection, training $\pi_\theta$
           to produce $y^{(2)}$ from $x$ only. \vspace{5pt}

\UNTIL{converged}

\end{algorithmic}
\end{algorithm}

We introduce \emph{Experiential Reinforcement Learning} (ERL), a training framework that enables a language model to iteratively improve its behavior through self-generated feedback and internalization. The key idea is to treat reflection as an intermediate reasoning signal that guides a refined second attempt, while reinforcement learning aligns both attempts with reward, and supervised distillation consolidates successful behaviors into the base policy. An overview is shown in \cref{fig:rlsr}, and the core training loop appears in \cref{alg:rlsr_simple}. A detailed implementation, including memory persistence and gating logic, is provided in Appendix \ref{sec:erl_details}.

Given an input task $x$, the model $\pi_\theta$ first produces an initial response
\[
y^{(1)} \sim \pi_\theta(\cdot \mid x),
\]
which is evaluated by the environment to produce textual feedback $f^{(1)}$ and reward $r^{(1)}$. Rather than immediately updating the policy, ERL optionally triggers a reflection-and-retry phase when the first attempt underperforms relative to a reward threshold $\tau$. This selective retry mechanism focuses compute on trajectories that are most likely to benefit from revision while avoiding unnecessary refinement when performance is already sufficient. When triggered, the model generates a reflection
\[
\Delta \sim \pi_\theta(\cdot \mid x, y^{(1)}, f^{(1)}, r^{(1)}, m),
\]
which serves as self-guidance describing how the initial attempt can be improved. Here, $m$ denotes a cross-episode reflection memory that persists successful corrective patterns discovered during training. This memory provides contextual priors that help stabilize reflection generation and encourage reuse of previously effective strategies. The model then produces a refined response
\[
y^{(2)} \sim \pi_\theta(\cdot \mid x, \Delta),
\]
and receives new feedback $(f^{(2)}, r^{(2)})$. Reflections that lead to sufficiently improved outcomes are stored back into memory, 
\[
m \leftarrow \Delta \quad \text{if} \quad r^{(2)} > \tau,
\]
allowing corrective knowledge to accumulate across training episodes. The reflection is assigned reward $\tilde r = r^{(2)}$, encouraging reflections that lead to improved downstream performance.

Both attempts and reflections are optimized using a reinforcement learning objective
\[
\mathcal{L}_{\text{policy}}(\theta)
=
- \mathbb{E}\!\left[
A \, \log \pi_\theta(y \mid x, \cdot)
\right],
\]
where $y$ denotes model outputs arising from the first attempt, reflection, or second attempt, and the conditioning context corresponds to the inputs specified in \cref{alg:rlsr_simple}. The advantage estimate $A$ is computed from the associated rewards.

While reflection and environment feedback provide strong training signals, such supervision is typically unavailable at deployment time, where the model must operate in a zero-shot setting. 
% \jz{maybe also add  sth like, when human learns, we also  `memorize'/`internalize' the experience? } 
We therefore introduce an internalization step that converts reflection-guided improvements into persistent policy behavior. The goal is to make the model remember corrections discovered during training and avoid repeating the same mistakes when feedback is absent. We implement internalization via selective distillation: we supervise the model to imitate only successful second attempts while removing reflection context from the input. Concretely, given a training example $x$, we generate a refined response $y^{(2)}$ and reward $r^{(2)}$, and optimize
\[
\mathcal{L}_{\text{distill}}(\theta)
=
- \mathbb{E}\Big[ \mathbb{I}\!\left(r^{(2)} > 0\right)\, \log \pi_\theta\!\left(y^{(2)} \mid x\right) \Big],
\]
where $\mathbb{I}(\cdot)$ is the indicator function. This trains $\pi_\theta$ to reproduce improved behavior from the original input $x$ alone (no reflection), ensuring that lessons learned through feedback and self-reflection persist at test time.

By alternating between reinforcement learning, selective reflection, and distillation, ERL bootstraps self-improvement: reflections guide higher-quality retries, memory preserves effective corrective structure, reinforcement learning aligns behavior with reward, and distillation internalizes gains into the core model. Over time, this interaction stabilizes training, concentrates exploration on failure cases, and reduces dependence on explicit reflection at inference.

\subsection{Comparison to Standard RLVR}

Standard reinforcement learning with verifiable rewards (RLVR) optimizes a policy directly from scalar outcome signals. Given an input $x$, the model samples a response $y \sim \pi_\theta(\cdot \mid x)$ and receives a reward $r$, with policy updates derived from trajectory-level credit assignment. In this formulation, feedback influences learning only through reward-driven optimization, requiring the model to implicitly discover how failures should translate into behavioral change. Corrective structure therefore emerges slowly through repeated exploration, with no explicit mechanism for revising behavior within the same learning episode. This learning dynamic corresponds to trial-and-error optimization, as illustrated in Figure~\ref{fig:rlsr_comp}.

Experiential Reinforcement Learning (ERL) augments this loop with an explicit experience–reflection–consolidation stage embedded inside each trajectory. Instead of optimizing solely from outcome reward, the model converts environment feedback into a reflection that conditions a refined attempt. This intermediate revision produces a locally improved trajectory that is reinforced and later internalized through selective distillation, allowing the base policy to reproduce corrected behavior without reflection at inference. A cross-episode reflection memory further stabilizes this process by preserving corrective patterns that proved effective, allowing subsequent reflections to reuse prior improvements. Importantly, ERL preserves the underlying RLVR objective: policy gradients remain reward-driven, but operate over a richer trajectory structure that includes explicit behavioral correction. This reframing shifts feedback from a scalar endpoint signal to a catalyst for immediate revision, reducing reliance on undirected exploration while maintaining compatibility with standard reinforcement learning pipelines. This contrast between blind trial-and-error learning and reflection-guided revision is visualized in Figure~\ref{fig:rlsr_simple} and Figure \ref{fig:rlsr_comp}.

\section{Experiment}
\label{sec:experiment}
We evaluate Experiential Reinforcement Learning (ERL) against standard RLVR on a set of agentic reasoning tasks.

\subsection{Task}
We evaluate ERL on three agentic reasoning tasks: Frozen Lake, Sokoban, and HotpotQA \citep{yang-etal-2018-hotpotqa}. Detailed environment descriptions are provided in Appendix~\ref{sec:env_details}.

For Frozen Lake and Sokoban, we configure the environments with sparse terminal rewards following \citet{wang2025cogitoergoludoagent} and \citet{guertler2025textarena}. The agent receives reward only at episode completion: a reward of +1 is assigned for successfully achieving the objective and 0 otherwise. Crucially, \textit{we do not provide explicit game rules or environment dynamics}. The model must infer task structure purely through interaction, with access limited to the available action set. This evaluation design is inspired by prior work on learning from experience, where the goal is to measure an agent’s ability to acquire task understanding through trial-and-error rather than relying on human-authored priors embedded in pretraining. The combination of sparse rewards and unknown dynamics therefore creates a challenging setting that emphasizes reasoning, planning, and experiential learning.

HotpotQA is adapted into an agentic multi-hop question-answering task following Search-R1 \citep{jin2025searchr}. Given a question, the model performs iterative tool-assisted retrieval before producing a final answer. To maintain consistency with the experiential learning setup, we provide only a default system prompt describing available tools, without additional task-specific guidance. Correctness is evaluated using token-level F1 against ground-truth answers. The reward function assigns 1.0 for exact matches, a proportional reward for partial matches with F1 score $\geq 0.3$, and 0 otherwise.

\subsection{Models and Baselines}
In our experiments, we train Olmo-3-7B-Instruct \citep{olmo2025olmo3} and Qwen3-4B-Instruct-2507 \citep{yang2025qwen3technicalreport} using both standard RLVR and our proposed ERL paradigm, with GRPO \citep{shao2024deepseekmathpushinglimitsmathematical} serving as the underlying policy-gradient optimizer in all cases. To ensure stable training, we adopt common reinforcement learning techniques such as clipping, KL regularization, and importance sampling. Notably, the internalization stage in ERL naturally involves off-policy data, which can introduce additional instability. We therefore apply the same stabilization techniques during this phase to maintain consistent optimization behavior. Additionally, because ERL requires two attempts per task along with an additional reflection step, we allocate 10 rollouts per task for RLVR and half as many per task per attempt for ERL to equalize the training compute per task across methods. Full hyperparameters and implementation details are provided in Appendix~\ref{sec:training_details}.

\section{Result and Discussion}

\begin{figure*}[t]
  \centering
  \includegraphics[width=\textwidth]{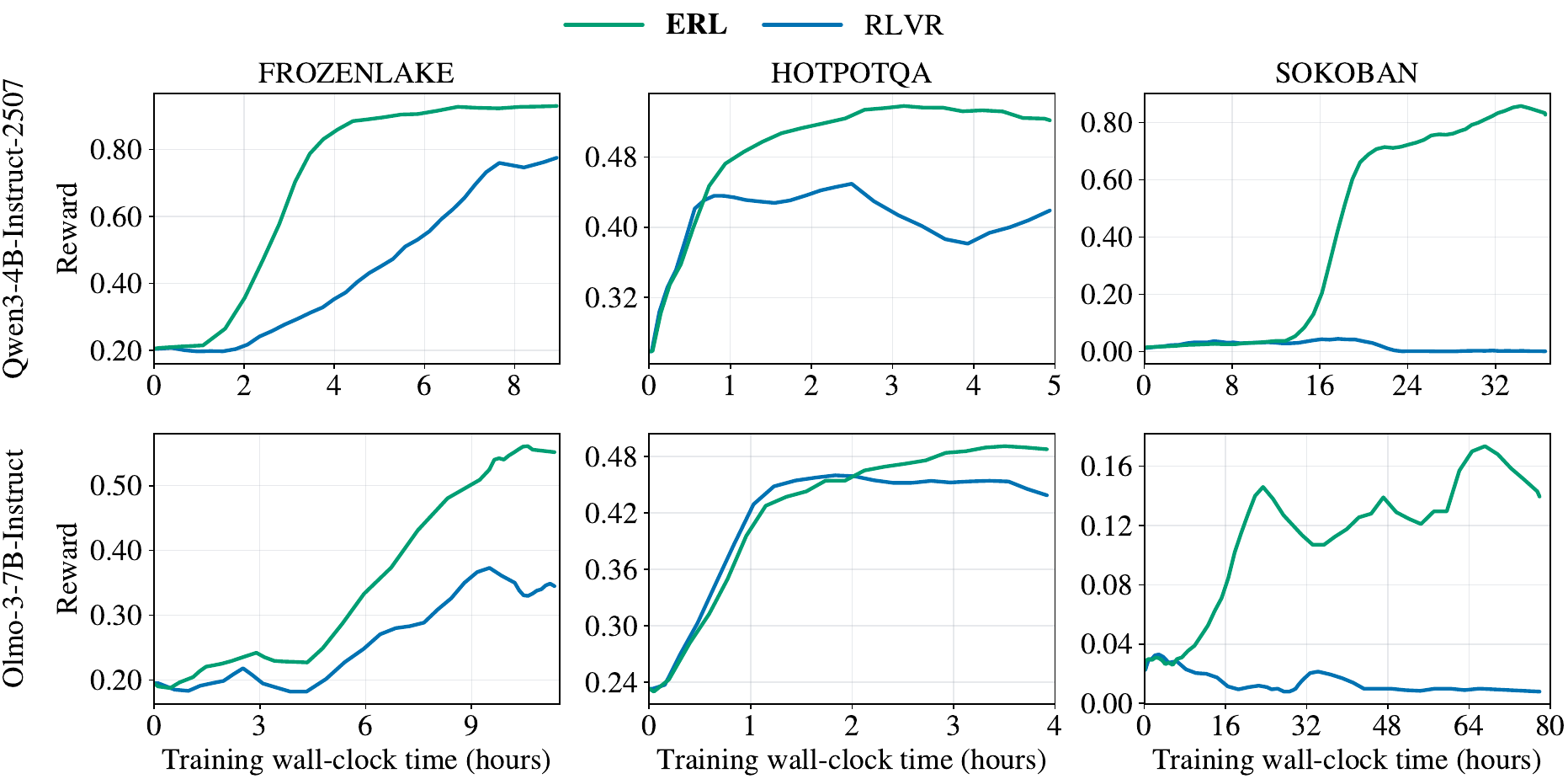}
  \vspace{-1em}
  \caption{Validation reward trajectories versus training wall-clock time on FrozenLake, HotpotQA, and Sokoban for Qwen3-4B-Instruct-2507 and Olmo-3-7B-Instruct. ERL consistently achieves higher reward and faster improvement than RLVR across tasks and models.}
  \label{fig:result_curves}
\end{figure*}

We evaluate Experiential Reinforcement Learning (ERL) against standard RLVR across three environments spanning sparse-reward control (FrozenLake, Sokoban) and agentic reasoning (HotpotQA). Table~\ref{tab:result} summarizes the final performance, while Figures~\ref{fig:result_bar}–\ref{fig:rlvr_pre_vs_post_reflection} visualize the performance and learning dynamics. All curves are smoothed with a trailing moving average over 5 points. The same smoothing procedure is applied to all figures unless otherwise noted.

\subsection{Performance Across Tasks}

\begin{figure*}[t]
    \centering
    \includegraphics[width=\linewidth]{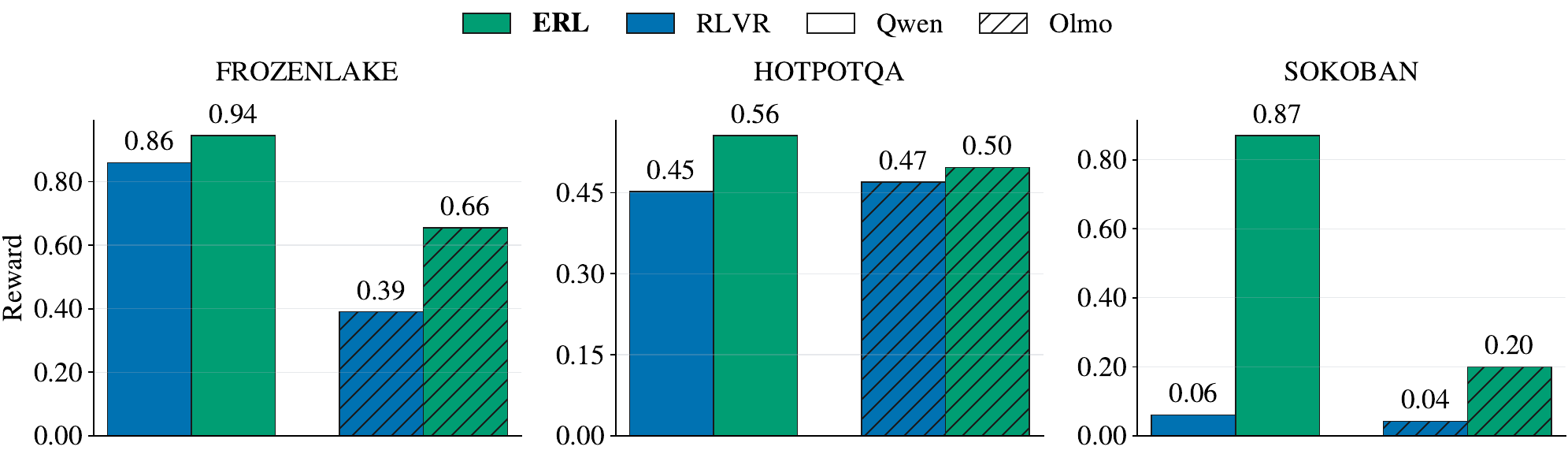}
    \vspace{-1em}
    \caption{Final evaluation reward on FrozenLake, HotpotQA, and Sokoban. ERL consistently outperforms RLVR for both Qwen3-4B-Instruct-2507 and Olmo-3-7B-Instruct.}
    \label{fig:result_bar}
\end{figure*}

ERL consistently improves final evaluation performance over RLVR across all tasks and both model backbones. As shown in Table~\ref{tab:result} and Figure~\ref{fig:result_bar}, experiential training yields gains ranging from moderate improvements on HotpotQA to substantial improvements on Sokoban and Frozenlake. 

The largest effect occurs in Sokoban, where Qwen3-4B-Instruct improves from 0.06 to 0.87 and Olmo3-7B-Instruct from 0.04 to 0.20. Sokoban requires long-horizon planning and recovery from compounding errors, making performance sensitive to how well the agent reasons about environment dynamics. Similarly, FrozenLake demands that the agent infer symbol semantics, action consequences, and terminal conditions purely through interaction under sparse rewards. Importantly, as described in Section~\ref{sec:experiment}, unlike many prior evaluation setups that provide explicit rules or environment structure, our environments expose only observations and action interfaces; the agent must infer task dynamics through trial-and-error. This design emphasizes learning from experience rather than relying on pre-specified priors, making structured revision particularly valuable. In these settings, the experience–reflection–consolidation loop enables the model to analyze failures, revise strategies, and internalize corrective behavior within each episode, producing large improvements in exploration efficiency and policy quality.

HotpotQA shows smaller but reliable gains. A likely explanation lies in differences in task structure. Compared to the grid-based control environments, HotpotQA presents a more homogeneous interaction pattern centered on repeated tool invocation and answer synthesis, with denser evaluation feedback and fewer latent dynamics to infer. Because RLVR already receives relatively informative gradients in this regime, the additional benefit of structured experiential revision is reduced. This contrast suggests that ERL yields the greatest advantage in environments where learning requires substantial reasoning about unknown dynamics and long-horizon consequences, rather than primarily optimizing over a stable interaction loop.

Importantly, improvements are observed across both models, indicating that the benefits of ERL arise from enhanced learning dynamics rather than architecture-specific effects.

% \begin{table}[t]
% \centering
% \renewcommand{\arraystretch}{1.2}
% \begin{tabular}{lcccc}
% \toprule
% \multirow{2}{*}{Task} &
% \multicolumn{2}{c}{\texttt{Qwen3-4B-Instruct}} &
% \multicolumn{2}{c}{\texttt{Olmo3-7B-Instruct}} \\
% \cmidrule(l){2-3} \cmidrule(l){4-5}
% & RLVR & $\mathbf{ERL}$ & RLVR & $\mathbf{ERL}$ \\
% \midrule
% FrozenLake & 0.86 & 0.96 \hlsecondarytab{(+0.10)} & 0.39 & 0.66 \hlsecondarytab{(+0.27)} \\
% HotpotQA     & 0.45 & 0.56 \hlsecondarytab{(+0.11)} & 0.47 & 0.50 \hlsecondarytab{(+0.03)} \\
% Sokoban    & 0.06 & 0.87 \hlsecondarytab{(+0.81)} & 0.04 & 0.20 \hlsecondarytab{(+0.16)} \\
% \bottomrule
% \end{tabular}%
% \caption{\label{tab:result}Final evaluation reward on FrozenLake, HotpotQA, and Sokoban. ERL consistently outperforms RLVR for both Qwen3-4B-Instruct-2507 and Olmo-3-7B-Instruct.}
% \end{table}

\subsection{Learning Efficiency and Optimization Dynamics}

Figure~\ref{fig:result_curves} compares validation reward against wall-clock training time. Across tasks and models, ERL reaches higher reward earlier and maintains a persistent margin over RLVR. This acceleration is especially pronounced in FrozenLake and Sokoban, where RLVR progresses gradually while ERL rapidly approaches high-reward behavior.

These dynamics suggest that reflection introduces an intermediate corrective signal that reshapes exploration. Instead of relying solely on terminal reward propagation, the model conditions on feedback and self-generated critique to revise its behavior. This concentrates training updates on trajectories that are already partially aligned with the objective, reducing inefficient exploration.

Even in HotpotQA, where rewards are denser and the environment is comparatively simpler, ERL maintains a consistent performance advantage over RLVR. Across environments, these results indicate that ERL achieves higher final reward while improving learning efficiency, demonstrating that structured experiential revision leads to faster and more effective policy improvement.

\subsection{Mechanistic Role of Reflection}

\begin{figure*}[t]
  \centering
  \includegraphics[width=\textwidth]{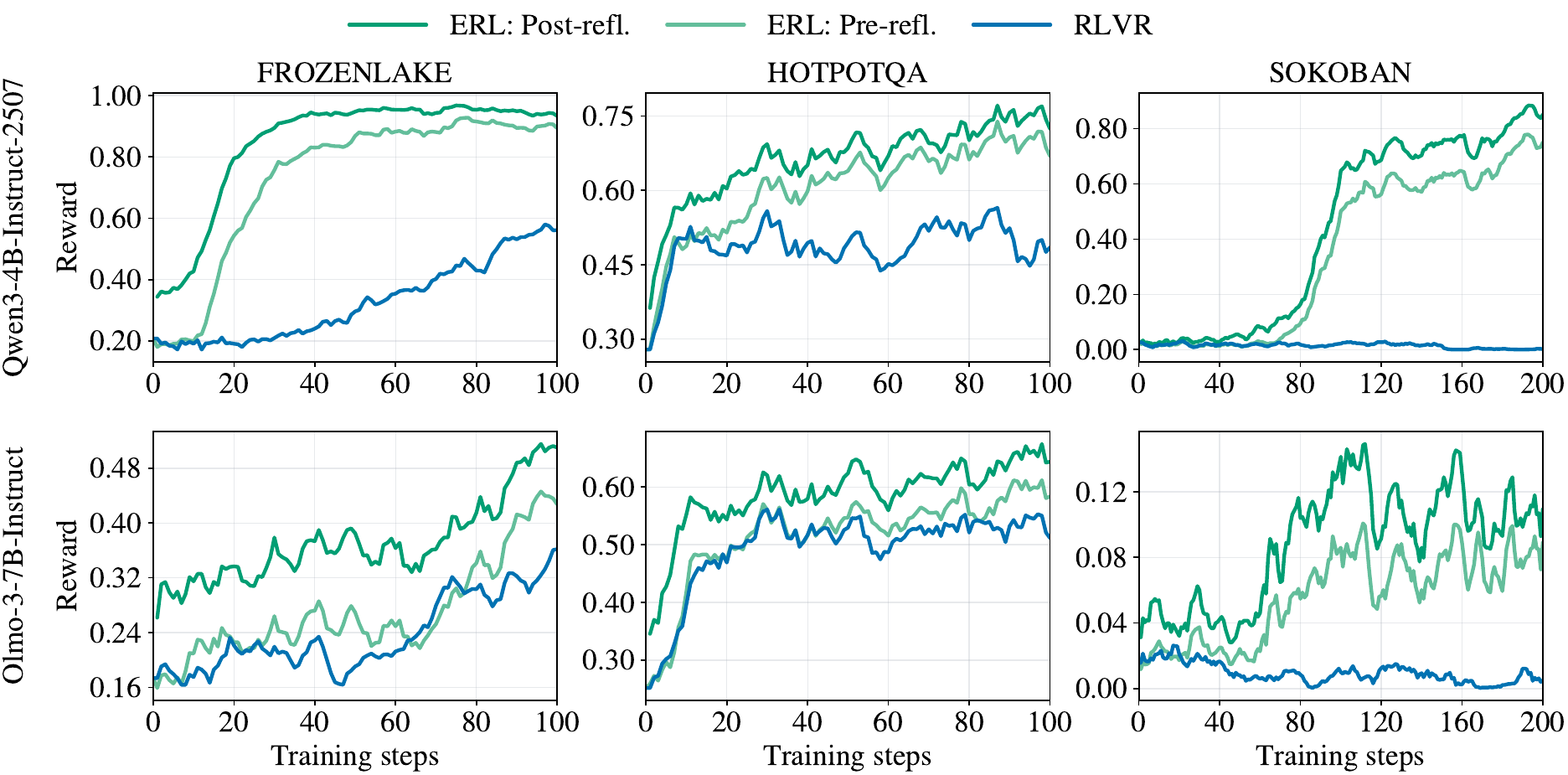}
  \vspace{-1em}
  \caption{Training reward trajectories for Qwen3-4B-Instruct-2507 and Olmo-3-7B-Instruct comparing RLVR with ERL before and after reflection. Post-reflection trajectories consistently achieve higher reward than both RLVR and pre-reflection trajectories.}
  \label{fig:rlvr_pre_vs_post_reflection}
\end{figure*}

Figure~\ref{fig:rlvr_pre_vs_post_reflection} shows training reward trajectories for ERL before and after the reflection step, alongside RLVR. Across environments and models, post-reflection trajectories consistently achieve higher training reward than pre-reflection trajectories and also exceed RLVR. 
%\jz{a qq: pos-reflection refers to generating the $y^{(2)}$?} \taiwei{yes}

This comparison highlights the immediate within-episode effect of reflection. After observing feedback from the first attempt, the model generates a structured revision that guides a second attempt with improved actions. The resulting gain in training reward indicates that reflection produces actionable corrections within the same episode, rather than only shaping behavior over long horizons. The sustained separation between pre- and post-reflection curves throughout training suggests that reflection serves as a systematic revision mechanism. By converting observed outcomes into targeted adjustments, it improves the quality of second attempts, which are subsequently reinforced and contribute to longer-term policy improvement.

\subsection{Ablation Study: Memory and Reflection Mechanisms}

\begin{table}[t]
\centering
\renewcommand{\arraystretch}{1.2}
\begin{tabular}{lcccc}
\toprule
Task & RLVR & ERL & ERL w/o Mem. & ERL w/o Refl. \\
\midrule

\addlinespace[3pt]
\multicolumn{5}{l}{\textbf{Qwen3-4B-Instruct-2507}}\\
\addlinespace[2pt]

FrozenLake & 0.86 & 0.94 & 0.86 \hlprimarytab{(-0.08)} & 0.60 \hlprimarytab{(-0.34)} \\
HotpotQA   & 0.45 & 0.56 & 0.56 \hlgraytab{(-0.00)} & 0.48 \hlprimarytab{(-0.08)} \\
Sokoban    & 0.06 & 0.87 & 0.87 \hlgraytab{(-0.00)} & 0.59 \hlprimarytab{(-0.28)} \\

\addlinespace[3pt]
\multicolumn{5}{l}{\textbf{Olmo3-7B-Instruct}}\\
\addlinespace[2pt]

FrozenLake & 0.39 & 0.66 & 0.64 \hlprimarytab{(-0.02)} & 0.54 \hlprimarytab{(-0.12)} \\
HotpotQA   & 0.47 & 0.50 & 0.47 \hlprimarytab{(-0.03)} & 0.46 \hlprimarytab{(-0.04)} \\
Sokoban    & 0.04 & 0.20 & 0.24 \hlsecondarytab{(+0.04)} & 0.06 \hlprimarytab{(-0.14)} \\

\bottomrule
\end{tabular}
\caption{Final evaluation reward on FrozenLake, HotpotQA, and Sokoban. ERL performance is compared against ablation variants, with highlighted drops showing the performance degradation relative to ERL when removing memory reuse (w/o Mem.) or structured reflection (w/o Refl.).}
\label{tab:result}
\end{table}

\begin{wrapfigure}{r}{0.5\linewidth}
  \centering
  \vspace{-0pt}
  \includegraphics[width=\linewidth]{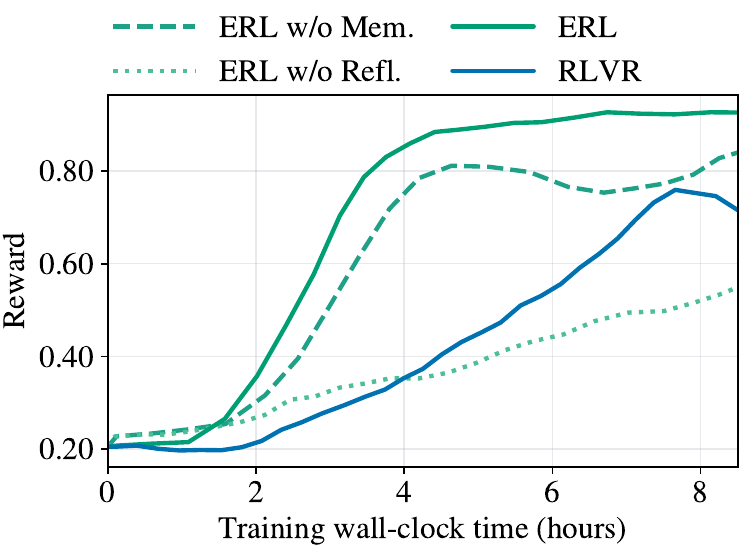}
  \vspace{-12pt}
  \caption{Ablation study on Qwen3-4B-Instruct-2507 in FrozenLake. We compare full ERL with two variants: (1) no memory, which disables cross-episode reflection reuse, and (2) no reflection, which replaces structured self-reflection with raw first-attempt context and a generic retry instruction.}
  \label{fig:ablation_curves}
  \vspace{-12pt}
\end{wrapfigure}

To understand how structured reflection and cross-episode memory contribute to performance, we conduct ablation studies across tasks and models. The quantitative results are reported in Table~\ref{tab:result}, and representative learning dynamics for FrozenLake with Qwen3-4B-Instruct-2507 are shown in Figure~\ref{fig:ablation_curves}. These experiments isolate individual components of ERL while keeping the overall training setup fixed.

The \textbf{no-memory} variant disables cross-episode reflection storage. Reflections are still generated and used to guide the second attempt within each episode, but they are not retained for reuse in future episodes. As a result, corrective signals remain local to individual trajectories rather than accumulating into persistent behavioral priors. 

The \textbf{no-reflection} variant preserves the two-attempt interaction structure but removes explicit structured reflection. Instead, the model receives the full first-attempt interaction history together with a generic instruction encouraging improvement. This design tests whether contextual reuse alone can replicate the benefits of structured reflective reasoning. The prompt template used in this setting is shown in Table~\ref{tab:generic_no_reflection_prompt} (Appendix).

The results in Table~\ref{tab:result} show a consistent ordering across most tasks and models: full ERL achieves the strongest performance, followed by the no-memory variant, while the no-reflection variant exhibits the largest degradation in most settings. Figure~\ref{fig:ablation_curves} further illustrates that removing memory slows convergence, whereas removing reflection substantially reduces both learning speed and final reward. These findings support the core design intuition of ERL: reflection generates actionable behavioral corrections, and memory propagates those corrections across episodes to enable cumulative refinement.

At the same time, Table~\ref{tab:result} reveals an important caveat. In the Olmo3-7B-Instruct Sokoban setting, the no-memory variant slightly outperforms full ERL. This suggests that when a model’s self-reflective ability is limited, or when the environment is complex and stochastic, persistent memory may propagate early inaccurate reflections, making recovery more difficult. In such cases, disabling cross-episode memory can mitigate the accumulation of erroneous priors. Nevertheless, across the broad set of tasks and models evaluated, ERL consistently delivers the strongest overall performance, demonstrating that structured reflection combined with persistent memory is highly effective in most practical settings.

\section{Related Work}
\paragraph{Reinforcement Learning for LLMs.}
Reinforcement learning has become a central approach for improving large language models. Early work focused on reinforcement learning from human feedback (RLHF) to align model behavior with human preferences and conversational objectives \citep{ouyang2022traininglanguagemodelsfollow, christiano2023deepreinforcementlearninghuman, shi2024saferinstructaligninglanguagemodels, shi2025wildfeedbackaligningllmsinsitu}. More recent efforts extend RL to enhance mathematical reasoning, where verifiable or programmatic rewards derived from executable checks or formal answer verification provide structured supervision for reasoning and solution construction \citep{openai2024openaio1card, Guo_2025, song2025hallucinationtaxreinforcementfinetuning, shi2026efficientreinforcementfinetuningadaptive}. In parallel, research on tool-using and agentic LLMs treats the model as a policy that interacts with external environments, alternating between actions and observations under task-dependent rewards to solve multi-step problems \citep{yao2023reactsynergizingreasoningacting, jin2025searchr, kimiteam2026kimik2openagentic, jiang2026modelrolesmultiturnmultiagent}. Despite their different goals, these approaches primarily treat environment feedback as a scalar optimization signal propagated through policy gradients, requiring the model to implicitly infer corrective structure through exploration. In contrast, our ERL paradigm introduces an explicit experience-reflection-consolidation loop that transforms environment feedback into structured behavioral revision before internalizing improvements into the base policy.

\paragraph{Learning from Experience.}
A growing body of work argues that the next scaling regime for AI will come not from more static human text, but from agents generating ever-richer data through interaction-i.e., learning predominantly from experience. \citet{SilverWelcomeTT} emphasizes that continual, agent-generated data streams and long-horizon decision-making as the route beyond imitation of human corpora. This motivates algorithmic mechanisms that convert failures into usable learning signal rather than relying on rare successes. In classic reinforcement learning, \citet{andrychowicz2018hindsightexperiencereplay} addresses sparse rewards by relabeling goals so that failed trajectories can still provide informative updates, substantially improving sample efficiency in goal-conditioned tasks. In the LLM-agent setting, \citet{zhang2025agentlearningearlyexperience} similarly targets the gap between imitation and full RL by training agents on their own interaction traces even when explicit rewards are unavailable, using the agent’s generated future states as supervision and including self-reflection as a way to learn from suboptimal actions. Meanwhile, inference-time reflection methods demonstrate that LLMs can critique and revise their own outputs to improve success \citep{zelikman2022star, madaan2023selfrefineiterativerefinementselffeedback, shinn2023reflexionlanguageagentsverbal}, but typically require reflection or memory at deployment.
Concurrent research explores integrating feedback-conditioned improvement directly into training. \citet{hübotter2026reinforcementlearningselfdistillation, song2026expandingcapabilitiesreinforcementlearning} formalize RL with textual feedback by distilling a feedback-conditioned teacher policy into a student policy. ERL is aligned with this direction but emphasizes explicit self-reflection as an intermediate reasoning step embedded inside the RL trajectory, where an initial attempt is followed by reflection and a refined retry. Coupled with selective internalization and cross-episode memory, this design treats reflection as a structured credit-assignment mechanism that transforms raw experience into durable behavioral improvement without requiring reflection at inference time.

\section{Conclusion}
In this work, we presented Experiential Reinforcement Learning (ERL), a training paradigm that incorporates an explicit experience–reflection–consolidation stage into the reinforcement learning loop to convert environment feedback into structured behavioral correction. By pairing reflection-guided revision with selective internalization, ERL enables models to learn corrective strategies during training and consolidate them into a deployable policy that operates without reflection at inference time. Across sparse-reward control and agentic reasoning tasks, ERL improves learning efficiency, stabilizes optimization, and produces stronger final policies relative to standard reinforcement learning baselines. These results demonstrate that embedding structured experiential revision directly into the training process provides an effective mechanism for translating feedback into durable behavioral improvement. Looking forward, this work suggests a path toward reinforcement learning systems that are fundamentally grounded in experience, where explicit reflection and consolidation become core primitives for building agents that continually learn, adapt, and improve from their own interactions.

\section*{Acknowledgements}
The authors thank the members of the LIME Lab and Microsoft Office of Applied Research for their helpful discussions, feedback, and resources.

\bibliography{references}
\bibliographystyle{colm2026_conference}

%%%%%%%%%%%%%%%%%%%%%%%%%%%%%%%%%%%%%%%%%%%%%%%%%%%%%%%%%%%%%%%%%%%%%%%%%%%%%%%
%%%%%%%%%%%%%%%%%%%%%%%%%%%%%%%%%%%%%%%%%%%%%%%%%%%%%%%%%%%%%%%%%%%%%%%%%%%%%%%
% APPENDIX
%%%%%%%%%%%%%%%%%%%%%%%%%%%%%%%%%%%%%%%%%%%%%%%%%%%%%%%%%%%%%%%%%%%%%%%%%%%%%%%
%%%%%%%%%%%%%%%%%%%%%%%%%%%%%%%%%%%%%%%%%%%%%%%%%%%%%%%%%%%%%%%%%%%%%%%%%%%%%%%
\newpage
\appendix
\onecolumn

\section{Full Algorithm and Gated Reflection}
\label{sec:erl_details}

\paragraph{Gated Reflection.}
Algorithm~\ref{alg:rlsr_full} presents the full version of ERL used in our experiments. Compared to the simplified version in Algorithm~\ref{alg:rlsr_simple}, the key difference is a gating mechanism on the second attempt: reflection and refinement are triggered only when the first-attempt reward satisfies $r^{(1)} < \tau$, where $\tau = 1$. In other words, reflection is applied only to failed or suboptimal trajectories.
In early experiments, we applied reflection to all trajectories, including successful ones, but this led to unstable training and reduced generalization. First, reflecting on already successful attempts encouraged reward hacking: the model sometimes generated instance-specific shortcuts that guaranteed success for the current sample but did not generalize to future episodes. Second, early in training when first-attempt rewards are typically low, the optimization signal became dominated by the second attempt and reflection, which are inherently off-policy relative to the base policy. This imbalance weakened the on-policy learning signal and destabilized the policy. The gating mechanism mitigates these issues by ensuring that successful trajectories remain purely on-policy, while reflection is reserved for corrective revision on failed attempts. This design also aligns training with deployment: at inference time, the model must generate $y \sim \pi_\theta(\cdot \mid x)$ without access to reflection $\Delta$ or feedback signals. By restricting reflection to corrective cases and preserving sufficient on-policy updates in every batch, the gating mechanism improves stability in training. 

\paragraph{Memory Extensions.}
Algorithm~\ref{alg:rlsr_full} also maintains a simple reflection memory that stores successful reflections as system prompt in plain text. A natural extension is to replace this mechanism with a more sophisticated agentic memory system. For example, before the reflection step (Alg. \ref{alg:rlsr_full}, Line 12), the model may retrieve relevant past reflections from a memory base conditioned on the current input $x$, and after a successful refinement, update the memory using a structured agentic memory update rule rather than direct overwrite. Such retrieval-and-update schemes would allow ERL to scale to more diverse and long-horizon tasks by enabling selective reuse and continual refinement of past corrective knowledge.

\paragraph{On-Policy Distillation.}
The internalization step in Algorithm~\ref{alg:rlsr_full} can also be generalized beyond supervised distillation. Instead of training $\pi_\theta$ to reproduce $y^{(2)}$ from $x$ using a standard distillation loss, one may adopt an on-policy reverse KL objective. Let the contextual policy with access to reflection and memory be $\pi_\theta(\cdot \mid x, \Delta)$, and the deployment policy be $\pi_\theta(\cdot \mid x)$. An on-policy distillation objective can be written as
\[
\mathcal{L}_{\text{OD}}(\theta)
:=
\mathbb{E}_{x \sim \mathcal{D}}
\Big[
\mathbb{I}\!\left(r^{(2)} > 0\right)\,
\mathbb{E}_{y \sim \pi_\theta(\cdot \mid x)}
\big[
\mathrm{KL}\!\left(
\pi_\theta(\cdot \mid x, \Delta)
\;\|\;
\pi_\theta(\cdot \mid x)
\right)
\big]
\Big].
\label{eq:on_policy_distill}
\]
which encourages the deployment policy to match the richer contextual policy while remaining on-policy with respect to $\pi_\theta(\cdot \mid x)$. This connects ERL to recent reverse-KL and on-policy distillation approaches \citep{agarwal2024onpolicydistill, hübotter2026reinforcementlearningselfdistillation, song2026expandingcapabilitiesreinforcementlearning} and provides a principled alternative to supervised internalization.

\begin{algorithm}[t]
\caption{Reinforcement Learning from Self-Reflection (Full Version)}
\label{alg:rlsr_full}
\small
\begin{algorithmic}[1]

\STATE \textbf{Inputs:} Language model $\pi_\theta$; dataset of questions $x$;
environment returning feedback $f$ and reward $r$, reward threshold $\tau$.
\STATE \textbf{Initialize:} reflection memory $m \leftarrow \emptyset$.

\REPEAT
\STATE Sample question $x$ from dataset.

\STATE \textcolor{softblue}{\textbf{// First attempt}}
\STATE Sample answer $y^{(1)} \sim \pi_\theta(\cdot \mid x)$.
\STATE Obtain feedback and reward $(f^{(1)}, r^{(1)})$.
\STATE \textcolor{softblue}{\textbf{// RL update on the first attempt}}
\STATE Update $\theta$ via $\mathcal{L}_{\text{policy}}(\theta)$ over the first attempt

\STATE \textcolor{softblue}{\textbf{// Gated second attempt}}
\IF{$r^{(1)} < \tau$}
    % \STATE \textcolor{softblue}{\textbf{// Retrieve memory}}
    % \STATE Retrieve relevant reflections
    %        $m \leftarrow \text{Retrieve}(\mathcal{M}, x)$.

    \STATE \textcolor{softblue}{\textbf{// Reflection with cross-episode memory}}
    \STATE Sample reflection \(\Delta \sim \pi_\theta(\cdot \mid x, y^{(1)}, f^{(1)}, r^{(1)}, m).\)

    \STATE Sample refined answer \(y^{(2)} \sim \pi_\theta(\cdot \mid x, \Delta).\)
    \STATE Obtain feedback and reward $(f^{(2)}, r^{(2)})$.
    \STATE Set reflection reward $\tilde r \leftarrow r^{(2)}$.

    \STATE \textcolor{softblue}{\textbf{// Store reflection only if improved beyond threshold}}
    \IF{$r^{(2)} > \tau$}
        \STATE Store reflection:
           $m \leftarrow \Delta$.
    \ENDIF

    \STATE \textcolor{softblue}{\textbf{// RL update on the second attempt}}
    \STATE Update $\theta$ via
       $\mathcal{L}_{\text{policy}}(\theta)$ over reflection and second attempt.

    \STATE \textcolor{softblue}{\textbf{// Internalization}}
    \STATE Update $\theta$ via $\mathcal{L}_{\text{distill}}(\theta)$
           to internalize reflection, training $\pi_\theta$
           to produce $y^{(2)}$ from $x$ only.
\ENDIF

\UNTIL{converged}

\end{algorithmic}
\end{algorithm}

\section{Envrionment Configuration Details}
\label{sec:env_details}

\subsection{Frozen Lake}

Frozen Lake is a grid-based navigation environment in which an agent must move from a start location to a goal location on an $n \times n$ grid. We configure our Frozenlake environment following a setup similar to those used in TextArena \citep{guertler2025textarena} and \citet{wang2025cogitoergoludoagent}. The grid size $n$ is sampled uniformly from $[2, 9]$. For each instance, the start and goal tiles are randomly selected as distinct positions. The grid layout is generated procedurally to ensure that at least one valid path exists between the start and goal.

Each non-goal tile is assigned as either a safe frozen tile or a hole according to a frozen-tile probability parameter $p$, sampled uniformly from $[0.6, 0.85)$. Holes represent terminal failure states, while frozen tiles are traversable. Transitions are deterministic: the agent’s chosen action directly determines its next grid position, subject to boundary constraints.

At every step, the agent observes a full textual representation of the grid. To reduce the influence of pretrained symbolic priors, we employ abstract symbols rather than semantically meaningful markers. The default encoding is:
\[
\texttt{A} = \text{agent position}, \quad
\texttt{B} = \text{goal tile}, \quad
\texttt{C} = \text{hole}, \quad
\texttt{D} = \text{safe frozen tile}.
\]
This representation encourages the model to infer environment dynamics through interaction rather than relying on prior associations. 

In addition to the textual presentation of the grid, the environment also appends structured textual feedback to the end of the interaction history after each action. This feedback communicates the outcome of the most recent transition and serves as the only explicit signal describing terminal or invalid events. The feedback messages are defined as follows:

\begin{itemize}
    \item \texttt{The agent reached the goal} — issued when the agent successfully enters the goal tile. The episode terminates with reward $1.0$.
    \item \texttt{The agent fell into the hole} — issued when the agent enters a hole tile. The episode terminates with reward $0.0$.
    \item \texttt{Hit the max step limit} — issued when the agent exhausts the fixed step budget. The episode terminates with reward $0.0$.
    \item \texttt{No valid actions were recorded.} — issued when the agent produces an invalid action or when the attempted action results in no state change, such as moving into a boundary. The episode continues unless the step budget is exhausted.
\end{itemize}

The default system prompt, self-reflection prompt, and example task are shown in Tables \ref{tab:prompt_frozenlake}, \ref{tab:updater_frozenlake}, and \ref{tab:example_frozenlake}.

The reward function is sparse. The agent receives a reward of $1.0$ if it reaches the goal tile and $0.0$ otherwise. Episodes terminate upon reaching the goal, entering a hole, or exhausting a fixed step budget of 8 actions.

For training, we generate 10,000 procedurally sampled instances. Evaluation is conducted on a disjoint set of 100 instances constructed using the same generation process.

\subsection{Sokoban}

Sokoban is a grid-based box-pushing environment in which an agent must place all boxes onto designated goal tiles. We configure our Sokoban environment following a setup similar to those used in TextArena \citep{guertler2025textarena} and \citet{wang2025cogitoergoludoagent}. Each instance is represented as an $n \times n$ grid, where $n$ is sampled uniformly from $[6, 8]$ in our procedural generator. We construct single-box, single-goal layouts with border walls, and randomly sample interior positions for the goal, box, and player, subject to non-overlap constraints.

To control difficulty, each generated layout is accepted only if its shortest valid solution is at most 8 moves (computed by BFS over player--box states). This guarantees solvability while keeping episodes short-horizon. Train and test splits are disjoint at the layout level.

At every step, the agent observes the full textual grid. As in FrozenLake, we use abstract symbols to reduce direct reliance on pretrained semantic priors. The default encoding is:
\[
\begin{aligned}
\texttt{A} &= \text{agent position}, \quad
\texttt{a} = \text{agent on box}, \quad
\texttt{B} = \text{box}, \quad
\texttt{b} = \text{box on goal}, \\
\texttt{C} &= \text{goal tile}, \quad
\texttt{E} = \text{wall}, \quad
\texttt{D} = \text{floor}.
\end{aligned}
\]

The action space is \{\texttt{Up}, \texttt{Down}, \texttt{Left}, \texttt{Right}\}. Moves are deterministic. The agent may push exactly one adjacent box only when the cell behind the box is free; it cannot pull boxes, move through walls, or move through boxes. Invalid moves produce no state change.

In addition to the grid observation, the interaction trace includes structured textual transition feedback after each action. The feedback messages are:

\begin{itemize}
    \item \texttt{The agent solved the puzzle (all boxes on goals).} — issued when all boxes are on goal tiles. The episode terminates with reward $1.0$.
    \item \texttt{The agent moved or pushed a box; puzzle not solved yet.} — issued when the action changes the state but the puzzle remains unsolved.
    \item \texttt{The agent did not move (likely hit a wall or tried to push into a blocked space).} — issued when the chosen move is ineffective (no state change).
    \item \texttt{Hit the max step limit} — issued when the fixed step budget is exhausted before solving. The episode terminates with reward $0.0$.
\end{itemize}

The default system prompt, self-reflection prompt, and example task are shown in Tables \ref{tab:prompt_frozenlake}, \ref{tab:updater_frozenlake}, and \ref{tab:example_sokoban}.

The reward is sparse: $1.0$ if and only if all boxes are on goals, and $0.0$ otherwise. Episodes terminate on success or when the step budget is exhausted. In the generated REEX Sokoban dataset, the per-instance step budget is 8.

For training, we generate 10,000 procedurally sampled instances. Evaluation is conducted on a disjoint set of 100 instances built with the same generation process.

\subsection{HotpotQA}

HotpotQA is a multi-hop open-domain question answering task in which an agent must answer compositional questions by retrieving and synthesizing evidence across multiple documents. Each instance consists of a natural-language question and a reference answer.

Unlike grid-based control environments such as FrozenLake or Sokoban, HotpotQA does not expose an explicit environment state. Instead, the agent operates through a tool-augmented interaction loop in which it alternates between reasoning, retrieval, and answer generation. The agent may invoke an external retrieval tool and ultimately produce a final textual answer. The solver instruction requires that the final answer be formatted inside \texttt{\textbackslash boxed\{\}} to enable reliable extraction.

The retrieval interface is defined as:
\[
\texttt{local\_search(query, top\_k)},
\]
which queries a local dense-retrieval server built over an indexed Wikipedia corpus and returns ranked text snippets relevant to the query. We use a Wikipedia corpus organized by \texttt{PeterJinGo/wiki-18-corpus}, with prebuilt dense indices from \texttt{PeterJinGo/wiki-18-e5-index}. Embeddings are generated using \texttt{intfloat/e5-base-v2}. Retrieval is powered by FAISS \citep{douze2024faiss} with multi-GPU support. During each episode, the agent is allowed up to 5 interaction turns, which may include reasoning steps, tool calls, and final answer submission.

Following the evaluation protocol of Search-R1 \citep{jin2025searchr}, the answer extracted from \texttt{\textbackslash boxed\{\}} is normalized prior to scoring by lowercasing and whitespace canonicalization. Correctness is measured using token-level F1 against the ground-truth answer. The reward function assigns a score of 1.0 for exact matches, a proportional reward equal to the F1 score when the F1 is at least 0.3, and 0 otherwise.

The default system prompt, self-reflection prompt, and example task are shown in Tables \ref{tab:prompt_hotpot}, \ref{tab:updater_hotpot}, and \ref{tab:example_hotpot}.

\begin{table*}[t]
\begin{tcolorbox}[colback=cyan!10!white]
{\bf Initial System Prompt for Frozenlake and Sokoban}

\tcblower

You are an agent playing a game on a grid, acting as a reasoning engine.\\
Your decisions are based on your current game rules (your best guess of how the game works) and your strategic playbook (your learned strategies). These may be incomplete or incorrect.\\
Your only way to interact with the environment is by choosing your NEXT ACTION.\\

\#\# Instructions\\
1. Analyze State: Summarize the current state.\\
2. Predict Long-term Value of Outcomes: Evaluate the strategic value and potential of the current state for the future.\\
3. Predict Immediate Consequences: For the top two candidate actions, predict their consequences using a ``result-because'' structure.\\
4. Select the Best Action: Choose the action leading to the most advantageous future state.\\

\#\# Required response structure\\
\texttt{<reason>}\\
\texttt{**1. Analysis of the Current State:** [Summary of the board state.]}\\
\texttt{**2. Prediction of the Value of Current States:** [Assessment] - Value: High / Medium / Low}\\
\texttt{**3. Prediction of Immediate Consequences:** [Top 2 candidate actions]}\\
\texttt{</reason>}\\

Then output the NEXT ACTION inside triple backticks, e.g., \texttt{```Up```}.\\

Always remember:\\
- Valid actions: Up, Down, Left, Right.\\
- Think step by step, but make the final line only the next action wrapped in triple backticks.\\

\end{tcolorbox}
\caption{\label{tab:prompt_frozenlake}Initial System used for Frozenlake and Sokoban.}
\end{table*}

\begin{table*}[t]
\begin{tcolorbox}[colback=cyan!10!white]
{\bf Initial System Prompt for HotpotQA}

\tcblower

You are a helpful assistant who answers questions directly and efficiently.\\

Provide your final answer in \texttt{\textbackslash boxed\{\}} format.\\

\#\# Available tool\\
\begin{verbatim}
[
  {
    "type": "function",
    "function": {
      "name": "local_search",
      "description": "Search for information using a dense retrieval
                      server with Wikipedia corpus",
      "parameters": {
        "type": "object",
        "properties": {
          "query": {
            "type": "string",
            "description": "Search query to retrieve relevant documents"
          },
          "top_k": {
            "type": "integer",
            "description": "Number of results to return (default: 5)",
            "minimum": 1,
            "maximum": 50
          }
        },
        "required": ["query"]
      }
    }
  }
]
\end{verbatim}

\end{tcolorbox}
\caption{\label{tab:prompt_hotpot}Initial system prompt used for HotpotQA.}
\end{table*}

\begin{table*}[t]
\begin{tcolorbox}[colback=cyan!10!white]
{\bf Self-reflection Prompt for Frozen Lake and Sokoban}

\tcblower

You are a chief scientific strategist and master tactician.\\
Your mission is to analyze extensive field data from numerous operations to distill and refine the Master Rulebook of a complex game.\\
You will be presented with a large collection of highly successful trajectories and critical failure trajectories, collected over a long period.\\
Your primary task is to perform a deep, comparative analysis to understand the fundamental principles of victory and defeat.\\
Act as a grand strategist, identifying universal patterns and high-level causal relationships.\\
Your goal is to synthesize these insights to produce the next generation's Master Rulebook, making it more robust, accurate, and effective.\\

Core Principles:\\
- Think Long-Term: focus on universal, strategic truths that hold across diverse scenarios.\\
- Learn from Contrast: extract insights by comparing winners and losers.\\
- Synthesize and Consolidate: produce a single unified theory.\\
- Be Authoritative and Concise: state rules as definitive principles.\\

Your output MUST be a single consolidated \texttt{<prompt>} block representing the new Master Rulebook:\\
\texttt{<prompt>}\\
\texttt{<game\_rules>}\\
\texttt{**1. Symbol Meanings:** [...]}\\
\texttt{**2. Information \& Interpretation:** [...]}\\
\texttt{**3. Gameplay \& Actions:** [...]}\\
\texttt{**4. Action Effects:** [...]}\\
\texttt{**5. Game Objective \& Termination:** [...]}\\
\texttt{</game\_rules>}\\
\texttt{<strategy>}\\
\texttt{**1. Core Strategies:** [...]}\\
\texttt{**2. Tactical Tips:** [...]}\\
\texttt{</strategy>}\\
\texttt{</prompt>}\\

\end{tcolorbox}
\caption{\label{tab:updater_frozenlake}Self-reflection prompt used for Frozen Lake and Sokoban.}
\end{table*}

\begin{table*}[t]
\begin{tcolorbox}[colback=cyan!10!white]
{\bf Self-reflection Prompt for HotpotQA}

\tcblower

You are an expert prompt updater.\\
You will analyze recent trajectories, tool calls, and rewards to improve the solver's system prompt.\\
When failures occur, explicitly add rules that prevent repeating them (e.g., missing tool calls, hallucinated facts, or unboxed final answers).\\
Keep the prompt short, actionable, and reusable.\\
Output ONLY the improved system prompt wrapped in \texttt{<prompt>...</prompt>} tags.\\

\end{tcolorbox}
\caption{\label{tab:updater_hotpot}Self-reflection prompt used for HotpotQA.}
\end{table*}

\begin{table*}[t]
\begin{tcolorbox}[colback=cyan!10!white]
{\bf Example Task for Frozen Lake}

\tcblower

\#\# \{System Prompt\} \\ \\
\texttt{Current Observation (0):}\\
\texttt{ D \;\; D \;\; C \;\; D }\\
\texttt{ A \;\; D \;\; D \;\; C }\\
\texttt{ D \;\; C \;\; D \;\; D }\\
\texttt{ D \;\; D \;\; B \;\; D }\\

You have not achieved the goal yet. Please give the next action.\\

\#\# Action space\\
\texttt{Up | Down | Left | Right}\\

\#\# Output requirement\\
Return reasoning in \texttt{<reason>...</reason>} and final action in triple backticks, e.g., \texttt{```Right```}.\\

\end{tcolorbox}
\caption{\label{tab:example_frozenlake}Example Frozen Lake task instance.}
\end{table*}

\begin{table*}[t]
\begin{tcolorbox}[colback=cyan!10!white]
{\bf Example Task for Sokoban}

\tcblower

\#\# \{System Prompt\} \\ \\
\texttt{Current Board (0):}\\
\texttt{ E \;\; E \;\; E \;\; E \;\; E \;\; E }\\
\texttt{ E \;\; A \;\; D \;\; B \;\; C \;\; E }\\
\texttt{ E \;\; D \;\; D \;\; D \;\; D \;\; E }\\
\texttt{ E \;\; E \;\; E \;\; E \;\; E \;\; E }\\

Puzzle not solved yet. Provide the next move.\\

\#\# Action space\\
\texttt{Up | Down | Left | Right}\\

\#\# Output requirement\\
Return reasoning in \texttt{<reason>...</reason>} and final action in triple backticks, e.g., \texttt{```Right```}.\\

\end{tcolorbox}
\caption{\label{tab:example_sokoban}Example Sokoban task instance.}
\end{table*}

\begin{table*}[t]
\begin{tcolorbox}[colback=cyan!10!white]
{\bf Example Task for HotpotQA}

\tcblower

\#\# \{System Prompt\} \\ \\
Question: \\
\textit{Which university did the author of ``The Hobbit'' attend?}\\

\end{tcolorbox}
\caption{\label{tab:example_hotpot}Example HotpotQA task instance.}
\end{table*}

\begin{table*}[t]
\begin{tcolorbox}[colback=cyan!10!white]
{\bf Second-Attempt Prompt Template for the No-Reflection Variant}

\tcblower

\#\# \{System Prompt\} \\ \\
You are also provided with the model's past attempt data, including observations, actions, rewards, and feedback.\\
Use this information as context to make a better next-attempt decision policy.\\
Follow the action/output format exactly.\\ \\
\{First Attempt's Trajectory\} \\

\end{tcolorbox}
\caption{\label{tab:generic_no_reflection_prompt}
Generic second-attempt system prompt used in the no-reflection ablation. The model is provided with the full first-attempt trajectory (observations, actions, rewards, and feedback) together with a generic instruction encouraging improvement, without any structured reflection signal.
}
\end{table*}

\section{Training Configuration Details}
\label{sec:training_details}
We train all models with the rLLM agent training stack \citep{rllm2025} using GRPO \citep{shao2024deepseekmathpushinglimitsmathematical}. Training runs on a single node with 8 H100s and uses vLLM \citep{kwon2023efficient} with FlashAttention \citep{dao2023flashattention2}. 

We enable hybrid engine training, gradient checkpointing, and remove-padding. The optimizer learning rate is \texttt{1e-6}. Actor updates use a mini batch size of 64, dynamic batch sizing, and a max token length per GPU of 24,000. FSDP parameter/optimizer offload is enabled for the actor, and parameter offload is enabled for the reference model.

We set the training batch size to 64, with a maximum prompt length of 8,196 tokens and a maximum response length of 8,196 tokens. Rollouts are generated asynchronously using vLLM in async mode with a tensor model parallel size of 1. We use a sampling temperature of 0.7, GPU memory utilization of 0.85. For validation rollouts, we generate 4 samples per prompt with temperature 0.7, top-p sampling set to 0.8, and top-k sampling set to 20.

KL regularization is enabled using a low-variance KL loss with coefficient 0.001, and we use a fixed KL control coefficient of 0.001. The actor clipping ratio upper bound is set to 0.28, and the entropy coefficient is set to 0. Rejection sampling and stepwise advantage estimation are disabled. 

For RLVR training, we generate 10 samples per prompt. For ERL training, we generate only 4 samples per prompt for each attempt to match the compute budget of RLVR. Evaluation is performed every 5 iterations, and training is manually early stopped upon convergence.

The design and implementation details of the ERL algorithm can be found in Appendix \ref{sec:erl_details}.

%%%%%%%%%%%%%%%%%%%%%%%%%%%%%%%%%%%%%%%%%%%%%%%%%%%%%%%%%%%%%%%%%%%%%%%%%%%%%%%
%%%%%%%%%%%%%%%%%%%%%%%%%%%%%%%%%%%%%%%%%%%%%%%%%%%%%%%%%%%%%%%%%%%%%%%%%%%%%%%

\end{document}